# Investigating the Common Authorship of Signaturesby Off-Line Automatic Signature Verification Without the Use of Reference Signatures

Moises Diaz , *Member, IEEE*,  Miguel  A. Ferrer , Soodamani Ramalingam ,and Richard Guest

**Abstract**

In automatic signature verification, questioned specimens are usually compared with reference signatures. In writer- dependent schemes, a number of reference signatures are required to build up the individual signer model while a writer-independent system requires a set of reference signatures from several signers to develop the model of the system. This paper addresses the problem of automatic signature verification when no reference signatures are available. The scenario we explore consists of a set of signatures, which could be signed by the same author or by multiple signers. As such, we discuss three methods which estimate automatically the common author- ship of a set of off-line signatures. The first method develops a score similarity matrix, worked out with the assistance of duplicated signatures; the second uses a feature-distance matrix for each pair of signatures; and the last method introduces pre-classification based on the complexity of each signature. Publicly available signatures were used in the experiments, which gave encouraging results. As a baseline for the performance obtained by our approaches, we carried out a visual Turing Test where forensic and non-forensic human volunteers, carrying out the same task, performed less well than the automatic schemes.

*Index Terms—* Off-line signature  verification,  biometrics, no reference signatures, feature-distance matrix, signature complexity.

## 1. INTRODUCTION

A Wide variety of enterprises store and use paper documents: delivery notes for transport companies, clinical histories of hospital patients [1], receipts, payments, banking transactions and legal documentation are just some examples of daily activities involving the digitization of documents. In the majority of cases, these documents include a personal handwritten signature.

It is often necessary to question whether the signatures on a set of stored signed documents belong to the claimed author of the signatures. Answering this question with any degree of certainty would imply having access to the stored reference signatures of the claimed author. Associating an ID to a set of stored signatures is not always possible. Among other reasons, asking for reference signatures may not be convenient, for example, in the case of VIP customers. Also, the storing of personal biometric data by a particular organization could be rejected by both staff and customers, because of the existence of laws or regulations preventing industrial operations from collecting and coding the signature specimens [2]–[4]. Similarly, reference signatures may not be available in dramatic scenarios such as terrorism cases, where signed extortion letters are manually examined [5] to decide whether the signatures are signed by the same person or not. Also, determining if there is a single serial killer or morethan one criminal involved can be determined by manual signature examination of signed notes left at the individual scenes of several murders [6] but obviously without reference signatures. The same applies in the case of other types of crime, including fraud [7], [8].

The unavailability of reference signatures led us to a completely new scenario where a particular stored set of signatures could be associated to the same author, without using any reference signature. An adequate solution to this issue would imply an advantage for the industry, which did not need to store reference signatures from their customers. In such a case, we ask the question: *could the automatic signature verification (ASV) field [9], [10] propose an automatic solution without reference signatures?*

Many developments in automated signature-based systems are often demonstrated in common benchmarks and international competitions. In recent signature competitions such as SigWiComp2013 [11], SigWIcomp2015 [12] or TSNCRV2018 [13], off-line Automatic Signature Verifiers (ASV) have been acknowledged to constitute scientific evidence. Typically, both industrial and academic researchers in such competitions evaluate their systems by solving challenging casework problems that require image-based or off-line signatures. As a baseline for our approach, a visual Turing Test is con- ducted with forensic and non-forensic volunteers. This allows us to expose the difficulty of the automatic task discussed in this paper with respect to human inspection.

Off-line ASVs tackle the dichotomous question of whether or not a particular signature belongs to a claimed person

[14], after a comparison with reference specimens based on two basic modes: writer-dependent (WD) and writer-independent (WI) signature verification (SV) systems [9], [10]. The WD-SV systems are based on a mathematical model for each enrolled writer [15]–[17], whereas, WI-SV systems develop a single model for all writers [18]–[20]. It is worth high-lighting that WI systems approach the forensic application model through computer vision and machine learning/pattern recognition strategies.

Regarding to WD-SV systems, the literature includes a number of proposals for off-line classification for genuine or non-genuine signature testing. Some examples are the Support Vector Machines (SVM) [21]–[23], Hidden Markov Models (HMM) [24], fuzzy member-ship functions [25], artificial immune systems (AIS) [26], distance-based classifiers [27], [28] or a combination of multiple classifiers based on dissimilarity score measures [29], among others [9], [17], [30].

On the other hand, WI-SV approaches commonly create a single model trained with a pair of genuine-genuine and genuine-forged specimens. These approaches can also create a dedicated classifier for each person by adapting from a universal background mode [20], [31]. A key advantage of the designed WI systems is that, once the statistical models are built, new users can be enrolled in the system and proceed with classification, without altering the operation of the system. Conversely, the same advantage is present when a previously enrolled user leaves the system.

In WI-SV, authors used to fix some thresholds in the ASV with a certain part of data and verify with another portion. Some proposals include fixing acceptance thresholds [32] or stability parameters with genuine samples [18], without training a classifier. Also, SVM and Multilayer Perceptron (MLP) classifiers are used in this modality [33] as well as SVM and a multi-layer neural network as bi-class classifiers, as in [19]. More recently, effective performances can be found by using deep learning methods in WI mode such a deep convolutional neural network in [34]. Another significant proposal involves using an ensemble of classifiers, as discussed in [35] and [36].

In addition, there are approaches which combine WI and WD schemes. Some contributions consider a challenging scenario with WI feature extraction techniques. These consist in learning a discriminant feature representation by using third party signatures, not enrolled in the evaluated system (e.g. [31], [34], [37]). This kind of hybrid system [37] was used later in a WD classification stage. Another hybrid system was effective in [38] where the authors extracted features through WD scores of the SigNet-F representation [34] and final results were obtained by a fusion of WI and WD classifiers at score level.

Nevertheless, previous works assume that during the verification of questioned signatures, there are typically five or ten reference signatures available [18], [25], [29]. The originality and central novelty of this work lies in the fact that we do not have reference signatures in any case studied here for verifying automatically the signatures. Instead, we have a set of two, three, four or five signatures and we estimate whether the same signer executed all of them. This implies a two-class classification problem: the signatures of the set belong to the same signer or do not belong to him or her.

*A. Contribution*

We presented a proof of concept of the idea of automatic signature verification without reference signatures at the 51st International Carnahan Conference on Security Technology (ICCST-2017) [39]. The method proposed in [39] built a squared matrix of similarity measures between the signatures of the set. Heuristically, a thresholding operation was set up to decide whether or not a set of signatures belonged to the same writer. Following comments and suggestions from conference delegates regarding the presentation of [39], and following collaboration with new researchers and two years of further work, the present paper proposes a significant step forward by deeply exploring the off-line automatic signature verification without the use of reference signatures.

Three competing methods are proposed in this paper to address this problem. While the first method is an improved version of the method introduced in [39], the other methods are newly introduced in this paper. Given a set of signatures whose common authorship is to be determined, the first method calculates a score similarity matrix with measures between the signatures of the set. A Least Square-Support Vector Machine (LS-SVM) [40] is used to decide whether or not the score similarity matrix belongs to the "all the signatures belong to the same signer" class. The second method employs feature-distance matrices for each pair of signatures in the set to be verified. Thus, if we had a set with four signatures, we would calculate six feature-distance matrices, one per pair. All these feature-distance matrices are combined, and an estimation with an LS-SVM [40] is taken with respect to common authorship. The third and last method is an extension of the second one, achieved by adding a complexity pre-classification of the signatures, which has a positive influence on the final decision.

The paper is broken down as follows. Section II describes the databases used, which were modified to create sets of signatures to evaluate their common authorship. Section III introduces the three proposed methods to evaluate the common authorship of a set of signatures without reference specimens. The experimental protocol is detailed in Section IV, while the experimental results of the different methods, as well as a visual Turing Test, are provided in Section V. Finally, Section VI concludes the paper.

## 2. DATASETS

In this paper, we evaluate our methods with four different datasets. Each dataset contains sets of signatures. Some sets comprise signatures belonging to the same author, whilst others contain signatures from different authors, i.e., genuine signatures and forgeries. These datasets of sets of signatures were built by selecting signatures from public corpuses of handwritten signatures. Specifically, we devised four data-bases:

**Dataset DS1:** Consists of sets of signatures selected from the GPDS-881 Off-line signature corpus [41]. The GPDS-881 corpus consists of 881 users with 21,144 genuine and 26,430 forgeries in total. The image-based signatures were scanned at 600 dpi with 256 grey levels and saved in PNG. This corpus was collected in the Group of Procesado De la Señal (GPDS) in a Spanish University, where specimens were acquired in a single session and each person used their own ballpoint pen.

**Dataset DS2:** Includes sets of signatures from the MCYT-75 Off-line corpus [42]. The MCYT-75 corpus includes 75 signers from four different Spanish universi- ties. The corpus includes 1,125 genuine and 1,125 deliberately forged signatures, acquired in two sessions. All the signatures were acquired with the same ink pen and the same paper templates. The paper templates were scanned at 600 dpi with 256 grey levels.

**Dataset DS3:** Is built with sets of signatures from the Dutch Off-line signatures, used in a signature competition organized during ICDAR 2009 [43]. Specifically, the available signatures consist of 12 authors, with 60 genuine signatures and 1,838 skilled forgeries in total. The signatures were saved in PNG format at 600 dpi with 256 gray levels. This corpus was collected by the Department Digital Technology and Biometrics at the Netherlands Forensic Institute (NFI).

**Dataset DS4:** This dataset selects signatures from the Dutch Off-line signatures SigComp 2011 [11]. A number of these signatures were also used in the signature competition organized in ICDAR 2009 [44]. There are 54 users available, with 1,292 genuine signatures and 639 forgeries in total. The original corpus was developed under supervision of forensic handwriting experts (FHEs) at the Netherlands Forensic Institute (NFI). Each image-based signature was collected with the same paper template and same ink pen. The images were saved as a PNG image at 400 dpi in colour format.

All the datasets, DS1, DS2, DS3 and DS4, were designed with the same structure. Each dataset contained 400 sets of signatures. To ensure a balanced study and proportional statistical results, these 400 sets were equally divided into sets with all the signatures of the same signer (genuine) and sets with signatures from different signers (genuine and forgeries). At the same time, the 400 sets were uniformly distributed into sets of two, three, four and five signatures. The distribution of sets per database is shown in Table I. The signer of each set, along with the genuine and forged samples used to make up the set, were randomly selected.

## 3. EVALUATING THE COMMON AUTHORSHIP OF A SET OF SIGNATURES

In the following, three methods for automatic off-line sig- nature verification without reference signatures are proposed. These are designed to tackle our central research question of whether or not all the signatures of a set belong to the same writer. The first method is based on a similarity measure between all the signatures of the set under study. It is an improved version of our past work [39]. The second method uses an ensemble of features and distance measures. This idea has been motivated by the effectiveness of ensemble strategies [35], [36], [45]. A possibility for improving the second method is by estimating the complexity of each signature and creating different models, according to their complexity category. Consequently, the third method is based on the same feature-distance matrices of method 2, along with the relevant complexity measures. Finally, from the matrices that com- pare the signatures, a Least Square Support Vector Machine (LS-SVM) is used to evaluate the common or non-common authorship of the set of signatures.

TABLE 1: STRUCTURE OF EACH DATASET USED IN THIS WORK. ALL SIGNATURES WERE RANDOMLY SELECTED

| # Signatures per set | # Sets with only genuine signatures | # Sets with genuine and forgeries | # Total sets |
|---|---|---|---|
| 2 | 50 | 50 | 100 |
| 3 | 50 | 50 | 100 |
| 4 | 50 | 50 | 100 |
| 5 | 50 | 50 | 100 |
| | | # Total sets | 400 |

A. Method 1: Similarity Score Matrix-Wise Method

This method compares all the signatures by means of a similarity measure, resulting in a square matrix. The similarity measure is obtained as a score produced by a statistical model as follows: 1) for each single signature of the set, a statistical model is built, and 2) a measure of similarity between each signature and the statistical model is carried out. The statistical model of each signature is built as follows: 1) the signature is duplicated, and 2) the statistical model is trained with the original and duplicated signatures.

To duplicate an off-line signature, a number of methods have recently emerged. One of the purposes of duplicating off-line signatures is to improve machine learning by introducing synthetic intra-personal variability. In the context of off-line signatures, several methods have been proposed to distort the images. Mapping equations to a signature image [46] or applying affine transformations [47] are the most popular techniques for broadening the number of training samples. In this work, we follow a cognitive-based distortion method described in [16] and [48]. This duplication model is inspired by the motor equivalence theory [49] and proceeds to execute an action such as handwriting. Apart from estimating the intra-personal variability, one of the advantages is that it requires only one signature as a seed, which is what we have in our case.

Formally, let $I_i$ be a signature $i$ in a set, $\forall i \in 1,\ldots,n$, where $n$ is the total number of signatures in a set. The signature $I_i$ is duplicated $m$ times, thus obtaining the duplicated signatures $\widehat{I_{i,c}}$, with $c \in 1,\ldots,m$. In our case, we chose $m = 20$. Hence, the model is trained with 21 signatures, which is a reasonable number to train a statistical model, as suggested in [16].

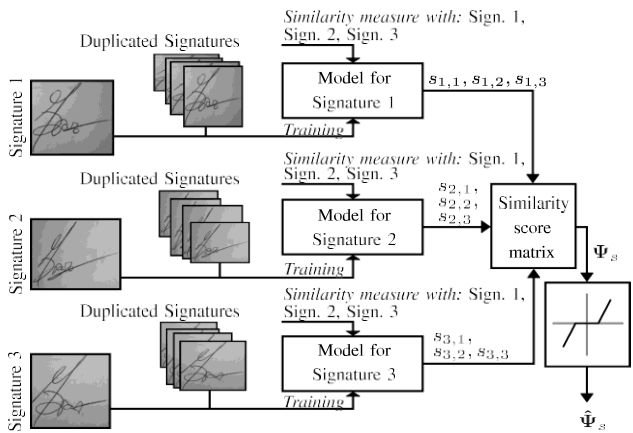

Fig. 1. Overview of method 1: similarity score matrix-wise method used to evaluate common authorship. An example is shown with a set of three signatures.

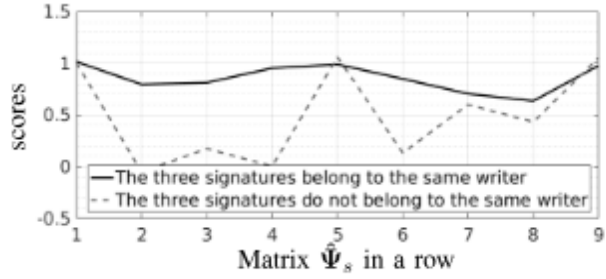

Fig. 2. *Example of similarity score matrices, $\widehat{\Psi_s}$, from two different sets of signatures. Matrix $\widehat{\Psi_s}$, are represented in a row for better visualisation.*

The statistical model of $I_i$ signature was worked out following [22], as follows: A basic version of local binary patterns (LBP) and local derivative patterns (LDP) were extracted for each signature $I_i$ and $\hat{I}_{i,c}$. These vector parameters were used to train the generative model $(M_i)$ based on a Least Square Support Vector Machine (LS-SVM) [40] with a radial basis function kernel [50]. The sigma and gamma hyper para- meters of the LS-SVM were calculated on the basis of a grid search and a two-fold cross validation strategy. The sigma value was searched on a grid of fifty values, logarithmically equally spaced between $10^0$ and $10^2$, whereas gamma was searched on a similar logarithmic sequence between $10^0$ and $10^3$. Once the model $(M_i)$ was built, we calculated the scores $s_{i,j}$ for each signature $(I_j)$ against the model $(M_i)$. This procedure was repeated for all $n$ signatures in the set.

The scores $s_{i,j}$ obtained denote the similarity measures. These scores are then used for designing a squared similarity score matrix of dimension $n \times n$, which we denote by $\widehat{\Psi_s}$. Figure 1 represents an example with a set of three signatures. The method generates three generative models, $M_i$, and, from each model, we test the initial signatures in the set. This leads to three output scores for each model, nine being the total number of scores $s_{i,j}$.

Some 95% of the scores $s_{i,j}$ are concentrated within the interval $(-1, +1)$, because of the use of the LS-SVM [40]. The closer to or further from $s_{i,j}$ to the limits of the interval, the clearer is supposed to be the classification. However, to facilitate the discrimination between common authorship and no common authorship classes, a non-linear transformation is applied to the similarity score matrix s, given as the result $\Phi_s$. Specifically, we apply a displacement to the scores $s_{i,j}$ in $\Psi_s$ by the following piecewise-defined function:

$$\widehat{S_{i,j}} = \begin{cases} s_{i,j} + R1 & if\ s_{i,j} > T1 \\ s_{i,j} - R2 & if\ s_{i,j} < T2 \\ s_{i,j} & \text{Otherwise} \end{cases} \quad \forall\ \{i \in (1,..., n), j \in (1,..., n)\} \quad (1)$$

where $s_{i,j}$ represents the transformed scores in $\widehat{\Psi_s}$, as is shown in Figure 1. Motivated by the output of the LS-SVM, the parameters used in this transformation were heuristically set as: R1 = R2 = 1 and T1 = T2 = 1. It is worthy highlighting that the transformation of Eq 1 has been adjusted for a LS-SVM classifier. However, such a transformation is also valid for classifiers that estimated posterior probability by adjusting the parameters R1, R2, T1, T2 as well as saturating the lowest and highest values at 0 and 1, respectively.

Similarity score matrices are shown to be effective in characterizing the common authorship of a set of off-line signatures. Figure 2 shows two similarity score matrices from two different sets. To allow a better visualization of the differences between them, matrices $\Psi_s$ are represented in a row. It is worth mentioning that the dimension of the matrix depends on the number of signatures. Thus, a set with n signatures results in a n × n matrix.

B. Method 2: Feature-Distance Matrix-Wise Method

In the previous method, the similarity between each signature pair is measured with a single score. This could be considered as being too simple a measure for such a complex problem. Method 2 improves the comparison between signatures of the set since several measures are worked out between each pair of signatures. Specifically, each signature $I_i$ is described by several different features $F_i^k$, k ∈ 1,..., K, with K being the index of the different features. To compare two signatures $I_i$ and $I_j$ within the same set, different distances d ∈ 1,..., D, are calculated between their features $F_i^k$ and $F_j^k$. Thus, for each pair of signatures, a feature-distance matrix of dimension K × D, called $FD_{ij}$, is built. Figure 3 shows an example of generated feature-distance matrices for a set with three signatures.

Generalizing, in a set of n off-line signatures, a mathematical combination of n signatures, taken in pairs (2 at a time), a set of $C_2^n = \binom{n}{2} = \frac{n!}{2!(n-2)!}$, different matrices is built. The dimension of each matrix is K × D, where in our case K = 10, represents the number of features used to describe the signature, and D = 15 is the distance between the features. The ten features used are detailed as follows:

- Six features based on geometrical features [51] are calculated. To that end, the signature is observed with both polar and Cartesian grid maps. Their lengths were fixed to 63 and 64 bins for polar and Cartesian features, respectively. The polar-based features divide the signature into normalised sectors. The difference between the two radii that define a sector, the angle of the sectors and the number of signature pixels in the sector are then computed. The Cartesian-based features utilise a superimposed grid. The distance from the centre to the envelope is then calculated for the horizontal and vertical features as well as for the signature transitions in the grid.

- Two textural features [22] are also employed. These are the local binary pattern (LBP) and the local derivative pattern (LDP). Each signature is divided into a 3 3 grid. Each region is overlapped to extract the LBP and LDP histograms. Both sequences of features are subsequently reduced with a discrete cosine transform to meet computational requirements, the LBP and LDP dimensions being 256 and 168 values, respectively.
Poset-oriented grid features [50]. The Equimass sampling grid method is employed on a thinned version of the signature. The dimension of this feature vector was 1280 bins. The features used are a representation of pixel transitions using lattice-shaped probing structures.

- Shape context [52]. The edge of the signature is calculated. A log-polar histogram with twelve bins for the angle and five bins for the radius is used. The number of pixels found in each bin is stored and then used as a feature vector of 256 values. Its use in signature verification has recently been demonstrated [53].

We then make independent comparisons for each particular type of feature. Such comparisons are performed using fifteen distances, described as follows:

- Dynamic Time Warping [54]. This distance represents the optimal elastic alignment between two sequences. The warping path is calculated in order to reduce the sum of Euclidean distances.

- Normalised Dynamic Time Warping. This is the distance obtained from the previous distance, divided by the warping path length.

- Minimum edit distance [55]. This dynamically calculates the minimum number of operations needed to convert one feature vector into another. The edit operations are deletion, insertion, and substitution of values into such a feature vector.

- Hungarian method [56]. The matching cost between two feature vectors is performed using the Chi-Squared distance. The Hungarian method utilises a subtraction technique to find the best assignment of elements that minimizes the total cost matrix.

- Eleven histogram similarities measurements [57]. Because of their simplicity and robustness in reporting statistical results when two feature vectors are matched, we used the following histogram-matching methods: intersection function, Chi Squared distance, Jeffrey Divergence, Kolmogorov-Smirnov distance, Hellinger distance, Bhattacharyya distance, $L_1$ or Manhattan norm and $L_2$ or Euclidean norm. Additionally, the $L_1$ and $L_2$ norm are applied to the cumulative sum of the feature vectors, which define how the data grows along the elements of the histograms. A pairwise distance in the form of a matrix is also calculated between the features obtained by means of a Chi-Squared distance. As proposed in [58], we store the best match as the minimum value of such a matrix.

Each comparison generates a feature-distance matrix $\mathbf{FD}_{ij}$, where $i$ and $j$ refer to signatures from the set, which can either come from the same or from different writers. Figure 4 shows an example of these two types of feature distance matrices. Their differences are better highlighted by representing the matrices as vectors. In contrast with Method 1, the dimensions of each matrix in this method are the same, independent of the number of signatures to be evaluated within the sets. We have a varying number of matrices, depending on the number of signatures in the set.

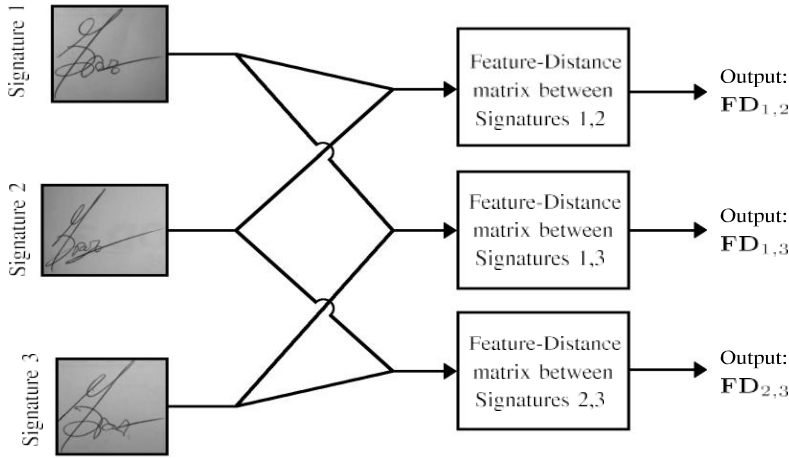

*Fig. 3. Illustration of method 2: matrix obtained with the feature-distance wise-method.*

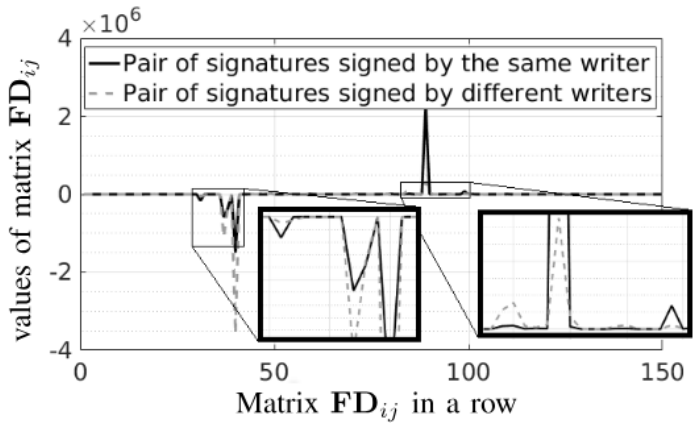

*Fig. 4. Example of two different Feature-Distance matrices for the comparison of two different pairs of signatures. Matrices FDij are represented in a row for a better visualisation.*

C. *Method 3: Feature-Distance Matrix with Complexity Method*

Method 2 is expected to improve on Method 1 because the comparison between signatures is more extensive. However, method 3 includes a new characteristic in the comparison: the complexity of the signature. It involves dividing the entire problem into several sub problems depending on the complexity of the signatures being questioned and developing different strategies for each complexity level. This proposal is reasonable, as it is well known that the quality of forgeries depends on the complexity of the signature, among other properties [9], [59]–[63].

TABLE II: COMPLEXITY FEATURES EXTRACTED FROM SIGNATURES

| Number | Feature | Description |
|---|---|---|
| $F_1$ | x size | length in pixels of signature bounding box |
| $F_2$ | y size | height in pixels of signature bounding box |
| $F_3$ | pixel percent | percentage of bounding box pixels that are inked (ink density) |
| $F_4$ | hole percent | percentage of bounding box pixels that are fully enclosed by ink |
| $F_5$ | number of components | number of independent ink objects within signature |
| $F_6$ | median column pixels | median number of ink pixels in columns within signature bounding box |
| $F_7$ | percent column empty | percentage of columns with no pixels within signature bounding box |
| $F_8$ | median row pixels | median number of pixels in row within signature bounding box |

For these reasons, the complexity of signatures has been exploited in signature verification over the last ten years because of its discriminating properties [9]. The complexity of a signature can be defined as the difficulty in falsifying a particular specimen [50], [64]. To estimate such level of difficulty, different procedures have been proposed. On the one hand, models developed by forensic document examiners' opinions have been adopted in the literature [59], [60]. Also, authors have pointed out that there are three theoretical relationships known as complexity theory in [65]. In [66] the authors propose to take into account the signature lengths, the number of pronounced directional changes in the line, or overwritings, the length of the pen-downs and complex pen patterns. However, quantifying the complexity in signatures remains an open challenge [50], although significant advances have been recently made for on-line signatures [61]– [63].

In this work, to evaluate issues of complexity, eight separate features that empirically reflect the human understanding of static complexity were extracted from all genuine signatures of DS1 - DS4. These features led to the evaluation of a particular comparison of specimens which, through selection, could lead to a more secure and robust evaluation of the common authorship of a set. These features are shown in Table II.

A k-means (k=3) [50], [60] clustering was applied to each combination of features, from single features to all eight features, resulting in 255 combinations. K-means was used to investigate the optimal selection and divisions using these features, with a complexity description applied retrospectively. Thus, k-means was performed on z-normalised raw-data scores derived from the features to ensure an unbiased clustering. This is because linearities are present across all variables. This was assessed using a scatterplot matrix across the eight features. Three metrics were used to evaluate each k-means clustering:

*Consistency:* the number of individual subjects that had all their signatures grouped in the same complexity cluster. good features (and combinations) would be those where samples from an individual signer were consistently assigned to the same group.

*Spread:* a measure of the evenness of the sample distribution across each of the 3 clusters. If a feature results in an even distribution of samples between the three complexity clusters, the feature ranks highly. If all samples were allocated to one cluster, with two single outliers forming the other two clusters, this feature/combination would result in a low ranking.

*Correlation:* the Spearman correlation between the k-means grouping and the raw feature data. This metric evaluates the ability of an ordinal feature to be mapped into a complexity grouping rank (for example, low complexity=group 1, high complexity=group 3). This is calculated by finding the ranked correlation between the assigned group and the raw feature data. A strong correlation is present when, using our example, low feature values

4are assigned to group 1 and high feature values are assigned to group 3. We take the absolute value of the calculated correlation coefficient to count as a negative correlation (i.e. when low feature values are assigned to group 3 and high feature values are assigned to group 1).

Every combination of features shown in Table II was systematically selected. For every combination of features, a meanmetric value for the feature set was separately calculated for each factor (Consistency, Spread and Correlation) using the scores across all test subjects.

The mean metric values were then separately ranked from the best performing feature set to the worst performing for each factor. As well as evaluating each factor individually, each of the three ranks for a particular feature vector was summed. A rank-of-summed-ranks was calculated to show the overall performance. In this ranking, a low number indicated good performance across all three metrics. Table III shows the winning feature vectors across three criteria: *a)* best single feature by consistency, *b)* best single feature by rank-of- summed-ranks, and *c)* best combination of features (either single or multiple) by rank-of-summed-ranks.

The feature pertaining to the percentage of columns that contained no ink pixels produced the best single feature cor- relation, however, the grouping consistency within signers and the spread of signatures within groups were not balanced for this feature. The median column pixel count (F6) gives the best single feature across all three factors when evaluating the rank of-summed-ranks. The best combination of multiple features by rank-of-summed-ranks uses x size (F1), the median number of pixels in columns (F6) and the median number of pixels in rows (F8).

By employing this lowest ranking to define the complex-ity group assignment for each signature, the groups can be generalized as follows:

Group 1: Low-medium *x* size $(F_1)$, low-medium amount of ink per column $(F_6)$ and, low-medium amount of ink per row $(F_8)$.
Group 2: Medium-high *x* size $(F_1)$, low-medium amount of ink per column $(F_6)$ and, low-medium amount of ink per row $(F_8)$.
Group 3: Medium-high x size $(F_1)$, medium-high amount of ink per column $(F_6)$ and, medium-high amount of ink per row $(F_8)$.

Therefore, we assign a complexity level, $C_\tau$, $\tau \in (1, 2, 3)$, to each signature individually. With such a complexity level, feature-distance matrices $FD_{ij}$ between two signatures can be found within the following six types: $C_{11}$, $C_{22}$, $C_{33}$, $C_{12}$, $C_{13}$, $C_{23}$. Regarding the second method, the feature-distance matrix, calculated with the complexity method, gives a three-rank level of complexity along with the corresponding feature-distance matrices. Figure 5 provides an overview of this method.

TABLE III: COMPLEXITY GROUPING ANALYS

| Criteria | Consistency Score (High = Best) | Spread Score (High = Best) | Correlation Score (High = Best) | Features | Rank of Ranks (Low = Best) |
|---|---|---|---|---|---|
| Single Feature Consistency | 624 | 2395 | 0.653 | $F_5$ | 67 |
| Single Feature Rank of Ranks | 609 | 1060 | 0.892 | $F_6$ | 2 |
| Winner - *Lowest Rank of Ranks* | 611 | 624 | 0.481 | $F_1, F_6, F_8$ | 1 |

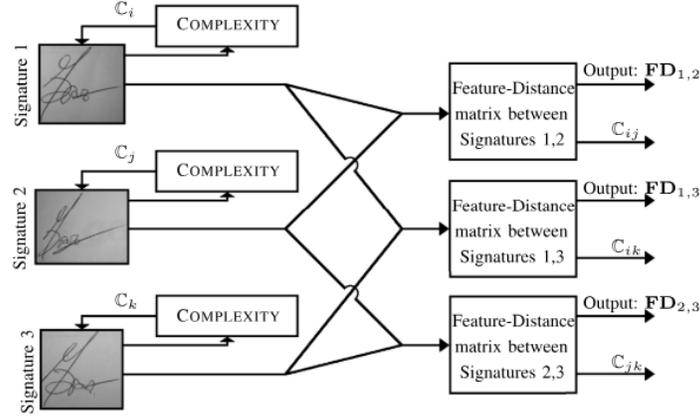

*Fig. 5. Overview of method 3: feature-distance matrix with complexity method.*

D. Estimating the Common Authorship of the Set of Signatures

The estimation of the common authorship of the set of signatures is similar in the three methods. In short, the above matrices with the measures of comparison between the signa- tures of the set are evaluated with a Least Square Support Vector Machine (LS-SVM) classifier. This gives an output score, which is used in deciding the common authorship of the set of signatures.

In Method 1, the simplest of the three, the similarity score matrix $\widehat{\Psi_s}$ is tested with an LS-SVM to obtain a classification score per set. As the dimension of the matrix depends on the number of users in the set, several LS-SVMs are trained: one for a set of 2 signatures, and another for a set of 3 signatures, and so on, up to the maximum number of signatures (five) considered in the sets.

In Method 2, we develop the mathematical combination of $C_n^2$ feature-distance matrices, $FD_{ij}$, in each set, with n being the number of signatures in the set. In this case, all the matrices are of the same dimension. Then, each $FD_{ij}$ is tested with an LS-SVM, and the LS-SVM being the same for all the matrices. In this way, we obtain $C_n^2$ scores. Several statistical combinations (minimum, maximum and average) of these scores are evaluated to get an output score of a set.

In Method 3, the procedure is similar to that of Method 2, but we train six LS-SVMs instead of just one. These six LS-SVMs are built by taking into account the six previously mentioned possibilities in terms of the three-rank complexity levels considered in this work. Accordingly, as each $FD_{ij}$ matrix has a corresponding complexity level, the proper LS-SVM is used for testing. Once we have the output score of the Cn 2 matrices of a set, the output score is obtained by the same statistical combination that we use in the second method.

The LS-SVMs are trained with one partition of the dataset. The details of the experimental protocol are given in the next section.

**4. EXPERIMENTAL PROTOCOL**

The objective of this section is to facilitate the replication of the experiments. To this end, we use four datasets: DS1, DS2, DS3 and DS4, as detailed in Section II. Each dataset contains 100 sets with two, three, four and five signatures for evaluation. In each case, 50 sets correspond to signatures executed by the same signer and 50 sets with signatures executed by more than one signer. Two partitions are made in the datasets: i) the sets for training, which use the half of the sets for training the LS-SVMs, and ii) the sets for testing, which include the second half of the dataset. Figure 6 illustrates the division of one of our databases for the experimentation.

In Method 1, the similarity score matrix-wise method, the dimensions of the matrices, $\widehat{\Psi_s}$., depend on the number of signatures to be evaluated. As such, we train the LS-SVM classifiers according to the signatures in the sets. For instance, the sets for training are used to train an LS-SVM for sets with two signatures. As positive samples, we use 25 matrices from sets of signatures executed by the same writer and, as negative samples, 25 matrices from sets of signatures executed by more than one writer. The same procedure is applied to sets with three, four and five signatures.

In total, we design four generative models. The remaining matrices generated in each testing partition are evaluated with the corresponding model. Upon completion, the output scores for the evaluation are stored in the corresponding class to evaluate the performance of the method.

In Method 2, the feature-distance matrix-wise method, we divide all sets of a particular dataset into sets for training and testing, independently of the number of signatures in the sets. It leads to $50 \times 4 = 200$ sets in each partition. The main reason is that the dimensions of all feature-distance matrices, $FD_{ij}$, are the same, i.e. $15 \times 10$. As such, we train a single LS-SVM with the feature-distance matrices from the training partition. As positive samples, we use $FD_{ij}$ from the pairs than involve signatures from the same signer. Conversely, as negative samples, we use matrices $FD_{ij}$ from the pairs that involve two different signers. In the testing partition, we obtain a score per set of signatures tested. Le t $C_n^2$ be number of feature-distance matrices to test in a set. We then obtain Cn 2 scores after the classification. We study several statistical combinations (minimum, maximum and average) of these scores to get an output score of the set.

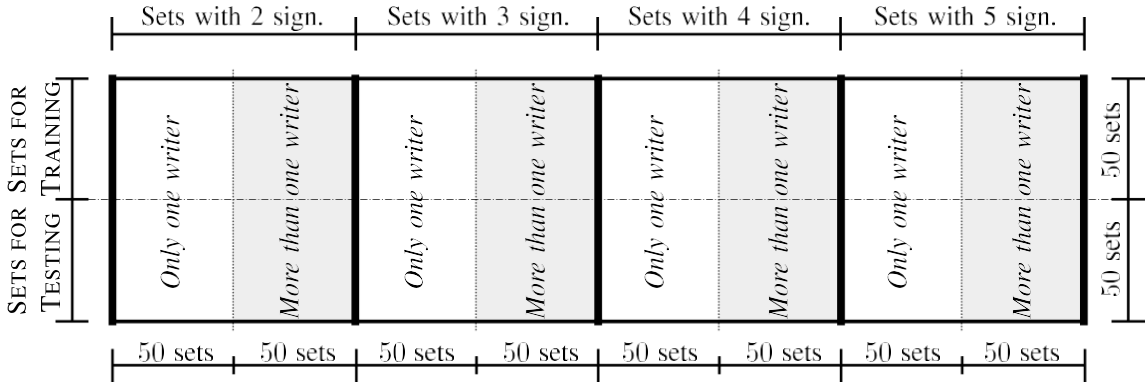

*Fig. 6. Dataset division into training and testing partitions for the experiments. This division is applied to all datasets (i.e., DS1, DS2, DS3 and DS4).*

In Method 3, the feature-distance matrix with complexity method, once again we divide all sets in the two partitions mentioned above, with 200 sets in each partition. Similarly, to the evaluation of the second method, the LS-SVM is trained with feature-distance matrices FDij obtained from pairs of signatures. As the third method provides the complexity of the feature-distance matrices, we divide all feature-distance matrices of the training partition into six groups, which correspond to the six cases of complexities considered in this work: C11, C22, C33, C12, C13, C23. Thus, we train six LS-SVMs, according to the complexity of the feature-distance matrices, $FD_{ij}$. Then, all 200 sets of signatures in the testing partition are classified with the corresponding LS-SVM model regarding the complexity of pairs of signatures. The output scores of the sets are obtained by the same statistical combination that we use in the second method. In our case, all LS-SVMs perform a grid-search on the hyperparameters in the ten-fold cross-validation for selecting the parameters in the sets included in the training partition [22]. The parameter settings that produce the best cross-validation accuracy are used in each case. The performance of all methods is evaluated using Detection error trade-off (DET) graphs for each dataset, DS1, DS2, DS3 and DS4. To build the DET graphs, we calculate the False Acceptance Rate (FAR) and the False Rejection Rate (FRR). The former indicates the error in classifying a set as executed by more than one person, i.e., genuine, and forged signatures, whereas the latter represents the error in classifying a set as executed by a single person, i.e., with only genuine signatures.

To quantify the error, we use the Equal Error Rate (EER) and the Area Under Curve (AUC) metrics. Each experiment was repeated ten times, after randomly choosing sets in the training and testing partitions. These experiments were performed independently on each database.

# 5. EXPERIMENTAL RESULTS

We aimed to validate the effectiveness and efficiency of automatic evaluation with respect to the common authorship of a reduced set of signatures. As such, the experiments were conducted with the aim of addressing two parallel outcomes. On the one hand, the experimental performance of the three proposed methods was analysed. On the other hand, the human capacity to decide if there is a single writer or more than one writer for a set of signatures was also evaluated as a baseline for comparison.

*A. Experiment 1: Evaluation of the Similarity Score Matrix-Wise Method (Method 1)*

This experiment allows evaluating the use of Support Vector Machines with the similarity score matrices, which is one of the improvements with respect to our previous work [39].

The experimental results are illustrated in Figure 7. They are similar in all cases, when considering both EER and AUC. Roughly, they are in the 21.20 24.40 range for EER and in the 82.17 88.41 range for AUC. In addition, to contextualise them, Table IV shows a comparison of DS1 and DS2 with our previous work presented in [39]. The same datasets, DS1 and DS2, were used in both the previous paper and in this work. The signature assessment in this work was carried out similarly to our previous work, which was presented in [39], with the main difference being the thresholding decision. In [39], the threshold for deciding if a set of signatures was from the same writer or from more than one writer was set heuristically at 0.1.

In contrast, in this work, the LS-SVM models were trained with 50 % of sets, equally distributed in sets with only genuine signatures and sets comprising both genuine signatures and forgeries. The remaining 50 % of the sets were then used for testing. Overall, we can conclude that this work is an improvement over the results obtained in our previous work.

TABLE IV: COMPARISON RESULT OF METHOD 1 WITH PREVIOUS WORK

| Method | Classification | DS1 | | DS2 | |
|---|---|---|---|---|---|
| | | EER | AUC | EER | AUC |
| *Previous work:* Duplicated Signatures + Similarity Score matrix [39] | Heuristically thresholding decision | 27.02 | 78.81 | 25.21 | 81.40 |
| *This work:* Duplicated Signatures + Similarity Score matrix | LS-SVM models | 22.30 | 83.71 | 18.50 | 88.41 |

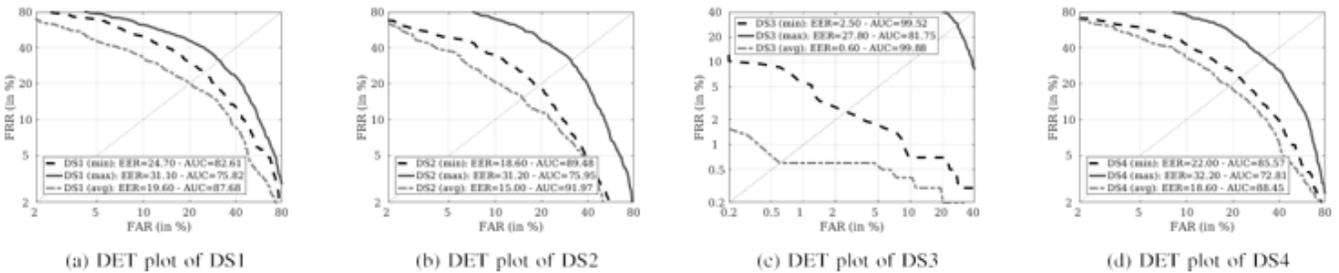

(a) DET plot of DS1   (b) DET plot of DS2   (c) DET plot of DS3   (d) DET plot of DS4

*Fig. 7. DET graph of Method 1 with DS1, DS2, DS3, DS4.*

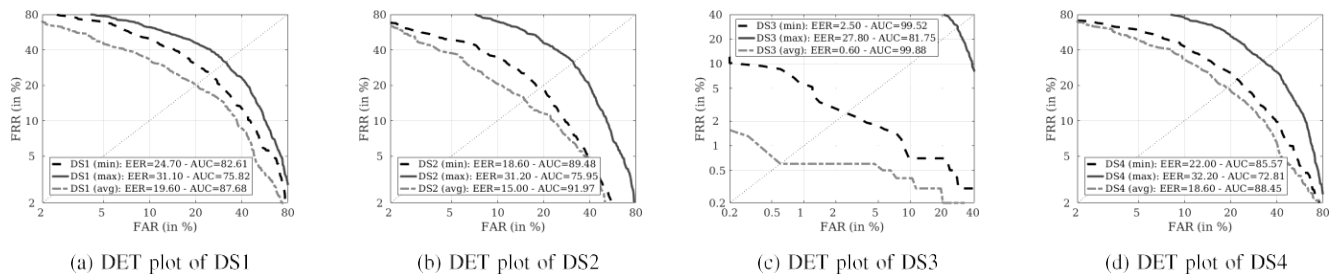

*Fig. 8. DET graph of Method 2 with all datasets. The score obtained from each individual set is calculated by the maximum, minimum and average of the $C_2^n$ scores of each set. (a) DET plot of DS1. (b) DET plot of DS2. (c) DET plot of DS3. (d) DET plot of DS4.*

*B. Experiment 2: Evaluation of the Feature-Distance Matrix-Wise Method (Method 2)*

The performances obtained for the feature-distance matrix- wise method are shown in Figure 8 for the four datasets. The first observation regards the statistic used for fusing the scores of each set. These results suggest that the most accurate estimator in all cases for fusing the scores is the average. It can be seen that, also in all cases, quantifying the average of the scores produces a result that is more robust than the minimum or maximum. Averaging the scores accounts for possible outlier scores from the individual signature comparisons. In addition to the EER, the AUC is provided. Regarding the performance of DS1 and DS2, it can be seen that the current results outperform those obtained with the previous method and in the previous research. In contrast, the datasets DS3 and DS4 have different behaviors. DS4 reports performance in line with DS1 and DS2, which suggests that this method is independent of the database type. However, DS3 reports a very competitive performance when the average estimator is used.

By observing its corresponding DET graph, with between 0.6% and 5.0% of error when only one writer should be recognised, we can see that the error in FAR is constant and is approximately 0.6%. This effect can be simply explained and is due to the fact that only 60 genuine signatures were available to design DS3, which reduced the variety of sets designed with signatures executed by the same writer. Finally, the results obtained lead us to conclude that Method 2 is a slight improvement over Method 1 and, therefore, preferred for our evaluation.

*C. Experiment 3: Evaluation of Complexity Effects in the Feature-Distance Matrix-Wise Method (Method 3)*

Here, we study the effect of considering three complexity levels in off-line signatures. The statistical quantification of the individual set of signatures by the minimum, the maximum and the average of their $C_2^n$ scores is also analysed. Figure 9 shows the DET graphs for all datasets as well as the performance obtained in terms of ERR and AUC.

Once again, it can be seen that averaging the scores is by far the most accurate option to improve the final performances. This observation is consistent with all experiments. For instance, as is shown in Figure 9b, the results are a significant improvement over other methods when the scores of a set of signatures are averaged. Regarding the final performances, we notice that Method 3 outperforms all results obtained so far. The main difference here is the use of a pre-classification of the signatures according to the complexity level. This leads to a more accurate modelling of the feature-distance matrices used in the training partition and more robust results.

D. Experiment 4: Visual Turing Test

In order to both stablish a baseline and analyse the human performance in evaluating whether a set of signatures are written by the same person or written by more than one person, we designed a visual Turing Test [67] in a similar way to that used previously [68], [69]. Our test consisted of 20 sets of signatures: 10 sets with signatures written by the same person and 10 sets with signatures written by more than one person. The number of signatures in each set was randomly selected. Figure 10 shows an example of a set included in the visual Turing Test.

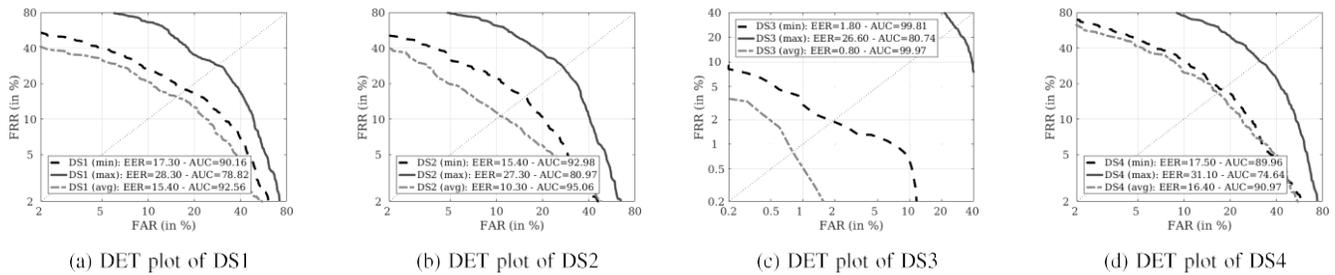

*Fig. 9. DET graph of Method 3 with all datasets. The score obtained from each individual set is calculated by the maximum, minimum and average of the $C_2^n$ scores of each set. (a) DET plot of DS1. (b) DET plot of DS2. (c) DET plot of DS3. (d) DET plot of DS4.*

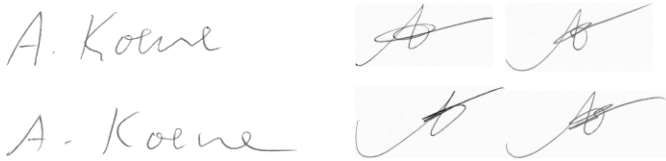

*Fig. 10. Example of set of signatures. (a) Set of two signatures written by the same person. (b) Set of four signatures written by more than one person.*

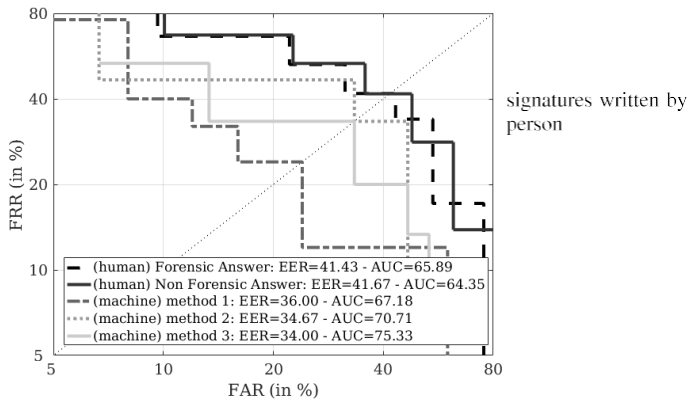

*Fig. 11. Performance of the human opinion through the visual Turing Test and the three proposed automated methods.*

In order to have more examples, three visual Turing Tests1 were designed with different sets in each test, but maintaining the distribution of ten sets written by the same writer and ten sets written by more than one writer. The order of the sets was randomly presented in the tests. Therefore, we had 60 sets to be evaluated in total. A total of 28 FHE and 301 non-FHE volunteers participated in the experiments, and in each test, we collected more than 2000 decisions in total. As an on-line survey was used, we collected responses from more than ten countries, including Ecuador, Spain, India, Italy, Colombia, Poland, Argentina and the UK. A seven-point Likert scale [70] was used to judge each set, where 1 meant that more than one writer executed the signatures in the set and 7 that a single writer produced the signatures. A response of 4 represented a confusion decision. Our method allowed us to calculate the False Same Signer Rate (FSSR) and the False Multiple Signers Rate (FMSR). FSSR means that the participant says that the same signer produced the set, but more than one signer actually produced it. FMSR denotes that participants believed that more than one signer produced the set, but only one signer actually produced it.

Additionally, the Average Classification Error (ACE) was calculated as ACE = (FSSR + FMSR)/2.

In order to statistically evaluate the three visual Turing Tests, a Kolmogorov-Smirnov and a Shapiro-Wilk test of normality distribution were used over the sequence of FSSRs, FMRSs and ACEs. As they were not normally distributed, a Kruskal-Wallis test, which is a non-parametric test, was carried out to evaluate the significant difference among them at a 0.05 significance level. We identified that the distribution of FSSR and FMRS across the three tests was the same (p > 0.05 in both cases). However, the null hypothesis was rejected when we analysed whether the distribution of ACEs was the same across the three tests (p = 0.002). In general, we can say that the complexity of each test was similar within the FSSR and FMRS results. We therefore, processed all the data together in order to obtain a larger population and meaningful statistical results.

In total, we obtained an FSSR of 42.32 %, an FMSR of 48.72%, and an ACE of 45.52%. It can be concluded that detecting whether the sets were produced by the same writer is a confusing task. It is worth noting that 50% represents a complete confusion. Moreover, an average response of 4.09 was received for sets with signatures written by the same writer and 3.57 for sets produced by more than one writer. This is another example of the confusion since 4 represented total confusion.

Further statistical tests were conducted to study the correlations between the age of the participants and ACE using a two-tailed Pearson test at a 0.01 level of significance. As the correlation coefficient was above 0, we observed that a correlation does not exist between these factors. As the machine-based experimental results were measured using EER and DET graphs, we also calculated these metrics for the visual Turing Test for forensic and non-forensic responses. Figure 11 shows these experimental results. We notice similar performances, 41.43% EER for an FHE response and 41.67% EER for a non-FHE response. Furthermore, there is no significant difference between FHE and non-FHE regarding accuracy, since the distribution of their FSSR, FMSR and ACE reported p > 0.2 applying both a Mann-Whitney U test and a Kruskal-Wallis test.

Additionally, the 60 sets validated by human opinion were evaluated with the three proposed methods. Once again, the most competitive performance was obtained with Method 3, with EER = 34.00% and AUC = 75.33%, as can be seen in Figure 11. As we are evaluating only 60 sets, the performances of the three machine-based methods are similar because of limited training and testing partitions found in the 60 sets. Although this reduces the statistical significance of the results, the performance obtained can validate the proposed methods against forensic and non-forensic opinion. Overall, in addition to once again highlighting the difficulty of the task, we notice a lower performance with the human decision than with the machine-based decision.

## 6. CONCLUSIONS

In this paper, we address a real-world problem that has not been previously studied in detail: that of verifying automatically off-line signatures without reference signatures. It leads to a new scenario in which it becomes possible to evaluate whether a set of off-line signatures belongs to the same signer or not. Among other examples, this problem can occur in the case of a series of related crimes in which signed notes are left by the perpetrator. Three novel methods are proposed to automatically answer this question. The first method consists of designing a similarity score matrix per signature set for evaluation. For populating this matrix, a signature duplicator [16] is used to enlarge the number of available signatures and to train as many generative models as possible as input signatures. In the second method, each signature is described by a vector of different features. A set of statistical distances is applied to these features, and a feature-distance matrix is generated per pair of signatures. This leads to a mathematical combination of feature-distance matrices from the signatures included in the set to be evaluated. The last method uses the concept of complexity of the signature to design more efficient feature-distance matrices. Three-ranked levels of complexity are used, which lead to a combination of six complexity cases for each set of two signatures under comparison. To evaluate the proposed methods, we use Least Square Support Vector Machine (LS-SVM) classifiers, which process either similarity score matrices or feature-distance matrices.

The key novelty of this paper versus prior literature in signature verification is that no reference signatures are available. For this reason, all three methods are configured by obtaining an LS-SVM model for each method. Once the models are established, the evaluation is carried out. We verify the proposed methods by random sets of signatures from four datasets of handwritten signatures. Additionally, the difficulty of executing this task for human examiners is demonstrated through a visual Turing Test as a baseline. We demonstrate our practical contribution to solving the problem since our automatic methods outperform the results obtained with the forensic and non-forensic human evaluation. This study, in our opinion, can be adapted to assist document analysts in similar tasks. Likewise, this work can open the door to the examination of new challenges in the field of biometric automatic signature verification as well as the redesigning of prior methods when no reference signatures are available.

**ACKNOWLEDGEMENT**

M. Diaz wishes to thank the Communication and Intelligent Systems Research Team of the University of Hertfordshire for hosting him during his postdoctoral visit in 2017, where this article was developed. We also thank to Elias N. Zois and Niclas Borlin for providing the poset-oriented features code and the Hungarian method, respectively.